\documentclass[11pt,a4paper]{article}
\usepackage[hyperref]{emnlp2020}
\usepackage{times}
\usepackage{latexsym}

\usepackage{microtype}
\usepackage{graphicx}
\usepackage{amsmath}
\usepackage{booktabs}
\usepackage{todonotes}
\aclfinalcopy 

\title{Inquisitive Question Generation for High Level Text Comprehension}

\author{Wei-Jen Ko$^1$\ \ \ \ 
Te-Yuan Chen$^2$\ \ \ \ 
Yiyan Huang$^2$\ \ \ \ 
Greg Durrett$^1$\ \ \ \ 
Junyi Jessy Li$^3$\\
$^1$ Department of Computer Science\\
$^2$ School of Information\\
$^3$ Department of Linguistics\\
The University of Texas at Austin\\

{\tt \{wjko, yuan0061, huang.yiyan\}@utexas.edu,}\\
{\tt gdurrett@cs.utexas.edu,jessy@austin.utexas.edu}

}

\date{}

\begin{document}
\maketitle
\begin{abstract}
Inquisitive probing questions come naturally to humans in a variety of settings, but is a challenging task for automatic systems. One natural type of question to ask tries to fill a gap in knowledge during text comprehension, like reading a news article: we might ask about background information, deeper reasons behind things occurring, or more. Despite recent progress with data-driven approaches, generating such questions is beyond the range of models trained on existing datasets.

We introduce {\sc Inquisitive}, a dataset of $\sim$19K questions that are elicited while a person is reading through a document. Compared to existing datasets, {\sc Inquisitive} questions target more towards high-level (semantic and discourse) comprehension of text. We show that readers engage in a series of pragmatic strategies to seek information. Finally, we evaluate question generation models based on GPT-2~\cite{gpt2} and show that our model is able to generate reasonable questions although the task is challenging, and highlight the importance of context to generate {\sc Inquisitive} questions.
\end{abstract}

\section{Introduction}

The ability to generate meaningful, inquisitive questions is natural to humans. Studies among children~\cite{jirout2011curiosity} showed that questions serving to better \emph{understand} natural language text are an organic reflection of curiosity, which ``arise from the perception of a gap in knowledge or understanding''~\cite{loewenstein1994psychology}. Because of its prominence in human cognition and behavior, being able to formulate the right question is highly sought after in intelligent systems, to reflect the ability to understand language, to gather new information, and to engage with users~\cite{vanderwende2007answering,vanderwende2008importance,piwek2012varieties,rus2010first,huang2017doesn}.
\begin{figure}[t]
\begin{center}
\includegraphics[width=0.48\textwidth,trim=0mm 0 0 0]{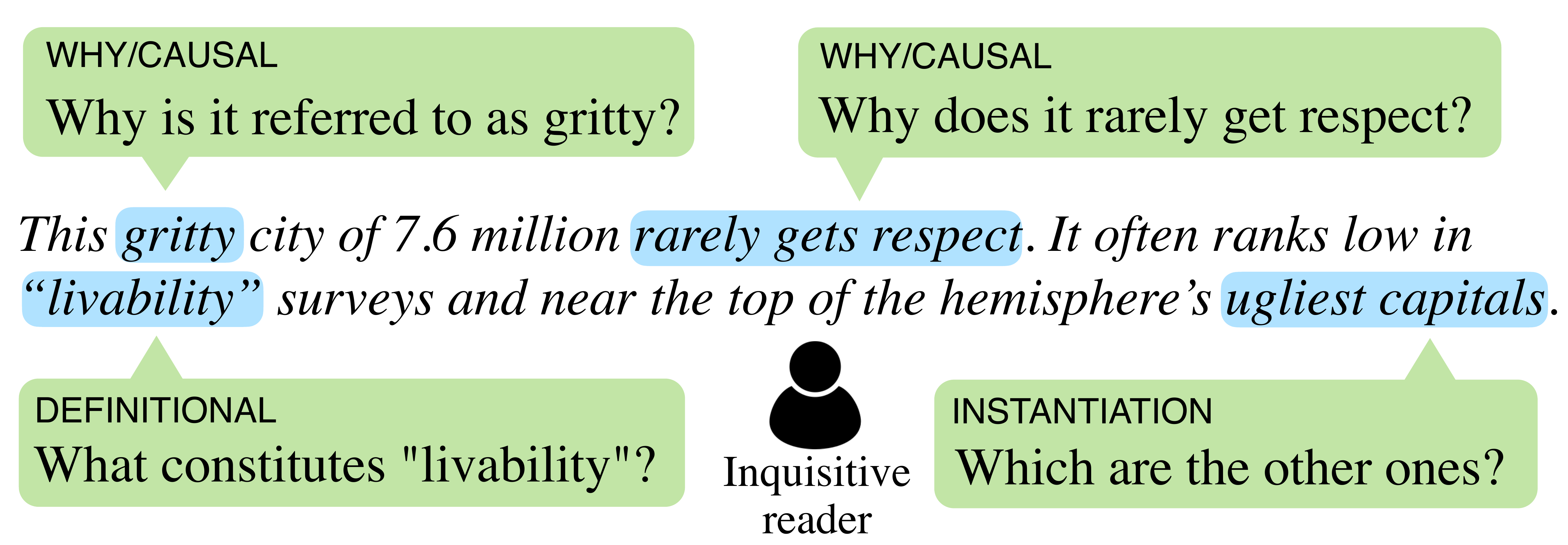}
\end{center}
\caption{Example questions in {\sc Inquisitive} reflecting a range of information-seeking strategies.}
\label{fig:questions}
\end{figure}
A recent line of work on data-driven question generation techniques~\cite{zhou,yuan2017machine,song2018leveraging,zhao2018paragraph} focuses on generating questions for datasets like SQuAD \cite{SQUAD,SQUAD2}.
However, factoid questions generated with an answer in mind after the user has read the full text look very different from more natural questions users might have \cite{NQ}. This has led to work on ``answer-agnostic'' question generation \cite{du1,Subramanian,scialom}, but the sources of data still emphasize simple factoid questions. Other prior work used question generation to acquire domain-specific knowledge~\cite{yang2018visual} and seek clarification in conversation~\cite{Clarification1,Clarification2,CQA}. However, data-driven generation of questions that reflect text understanding in a more general setting is challenging because of the lack of appropriate training data.

We introduce {\sc Inquisitive}, a new, large-scale dataset of questions that target high level processing of document content: we capture questions elicited from readers \emph{as they naturally read} through a document sentence by sentence. Because these questions are generated while the readers are processing the information, the questions directly communicate gaps between the reader's and writer's knowledge about the events described in the text, and are not necessarily answered in the document itself. This type of question reflects a real-world scenario: if one has questions during reading, some of them are answered by the text later on, the rest are not, but any of them would help further the reader's understanding at the particular point when they asked it.
This resource is a first step towards understanding the generation of such curiosity-driven questions by humans, and demonstrates how to communicate in a natural, inquisitive way by learning to rely on context.

Specifically, we crowdsource $\sim$19K questions across 1,500 documents\footnote{Data available at \url{https://github.com/wjko2/INQUISITIVE}}, each question accompanied by the specific span of the text the question is about, illustrated in Figure~\ref{fig:questions}. The questions are verified to ensure that they are grammatically correct, semantically plausible and meaningful, and not already answered in previous context. We show that the questions capture a variety of phenomena related to high-level semantic and discourse processes,
e.g., making causal inferences upon seeing an event or a description, being curious about more detailed information, seeking clarification, interpreting the scale of a gradable adjective~\cite{hatzivassiloglou2000effects}, seeking information of an underspecified event (where key participants are missing), and seeking background knowledge.
Our analyses reveal that the questions have a very different distribution from those in existing factoid and conversational question answering datasets \cite{SQUAD,newsqa,QuAC}, thus enabling research into generating natural, inquisitive questions.

We further present question generation models on this data using GPT-2~\cite{gpt2}, a state-of-the-art pre-trained language model often used in natural language generation tasks. Human evaluation reveals that our best model is able to generate high-quality questions, though still falls short of human-generated questions in terms of semantic validity, and the questions it generates are more often already answered. Additionally, our experiments explore the importance of model access to already-established common ground (article context), as well as annotations of which part of the text to ask about. Finally, transfer learning results from SQuAD 2.0~\cite{SQUAD2} show that generating inquisitive questions is a distinct task from question generation using factoid question answering datasets.

The capability for question generation models to simulate human-like curiosity and cognitive processing opens up a new realm of applications.
One example for this sort of question generation is guided text writing for either machines or humans: we could use these questions to identify important points of information that are not mentioned yet and should be included. In text simplification and summarization, these questions could be used to prioritize what information to keep. The spans and the questions could also help probing the specific and vague parts of the text, which can be useful in conversational AI. Because of the high level nature of our questions, this resource can also be useful for building education applications targeting reading comprehension.

\section{Related Work}

Among question generation settings, ours is most related to answer-agnostic, or answer-unaware question generation: generating a question from text without specifying the location of the answer~\cite{du1}. Recent work~\cite{du2,Subramanian,Wang,nakanishi2019towards} trains models that can extract phrases or sentences that are question-worthy, and uses this information to generate better questions. \citet{scialom} paired the question with other sentences in the article that do not contain the answers to construct curiosity-driven questions. However, these approaches are trained by re-purposing question-answering datasets that are factual~\cite{SQUAD} or conversational~\cite{QuAC,reddy2019coqa}. In contrast, we present a new dataset targeting questions that reflect the semantic and discourse processes during text comprehension.

Several other question answering datasets contain questions that are more information-seeking in nature. Some of them are collected from questions that users type in search engines \cite{wikiqa,MS,searchqa,NQ}. Others are collected given a small amount of information on a topic sentence \cite{newsqa,tydi} or in the context of a conversation \cite{QuAC,reddy2019coqa,qi}.
Our data is collected from news articles and our questions are precisely anchored to spans in the article, making our questions less open-ended than those in past datasets. While \citet{LREC} also collected a small number of reader questions from news articles, their goal was to study underspecified phrases in sentences when considered out-of-context.  Contemporaneously, \citet{tedq} presented a dataset of 2.4K naturally elicited questions on TED talks, with the goal to study linguistic theories of discourse.

Previous work generating clarification questions \cite{Clarification1,Clarification2,CQA} uses questions crawled on forums and product reviews. The answers to the questions were used in the models to improve the utility of the generated question. 
In our data, clarification is only one of the pragmatic goals. In addition, we focus on news articles which contains more narrative discourse and temporal progression.

\section{{\sc Inquisitive}: A corpus of questions}
This section presents {\sc Inquisitive}, a corpus of $\sim$19K questions for high level text understanding from news sources (Section~\ref{sec:data:sources}), which we crowdsource with a specific design to elicit questions as one reads (Section~\ref{sec:data:turk}). We then discuss a second validation step for each question we collected as quality control for the data (Section~\ref{sec:data:validation}).

\subsection{Text sources}\label{sec:data:sources}
In this work we focus on news articles as our source of documents. News articles consist of rich (yet consistent) linguistic structure around a targeted series of events, and are written to engage the readers, hence they are natural test beds for eliciting inquisitive questions that reflect high level processes.

We use 1500 news articles, 500 each from three sources: the Wall Street Journal portion of the Penn Treebank~\cite{ptb}, Associated Press articles from the TIPSTER corpus~\cite{ap}, and Newsela~\cite{newsela}, a commonly used source in text simplification (we use the most advanced reading level only).
We select articles that are not opinion pieces and contain more than 8 sentences to make sure that they are indeed news stories and that would involve sufficiently complex scenarios.

\subsection{Question collection}\label{sec:data:turk}
To capture questions that occur as one reads, we design a crowdsourcing task in which the annotators ask questions about what they are reading currently and without access to any upcoming context.

The annotators start from the beginning of an article, and are shown one sentence at a time in article order. After reading each sentence, they ask questions about the sentence, grounded in a particular text span within the sentence (via highlighting) that they would like elaboration or explanation of. We specifically asked for questions that would enhance their understanding of the overall story of the news article. An annotator can ask 0 to 3 questions per sentence, and the next sentence is only revealed when the annotator declares that no more question needs to be asked. The annotation instructions and interface is shown in Figures 4 and 5 in the Appendix.

In this manner, we elicit questions from annotators for the first 5 sentences from each article. We restrict the annotation to these sentences as they reflect a reader's cognitive process of context establishment from the very beginning, and that lead sentences are known to be the most critical for news articles~\cite{errico}. 
\begin{figure}
\begin{center}
\includegraphics[width=0.48\textwidth]{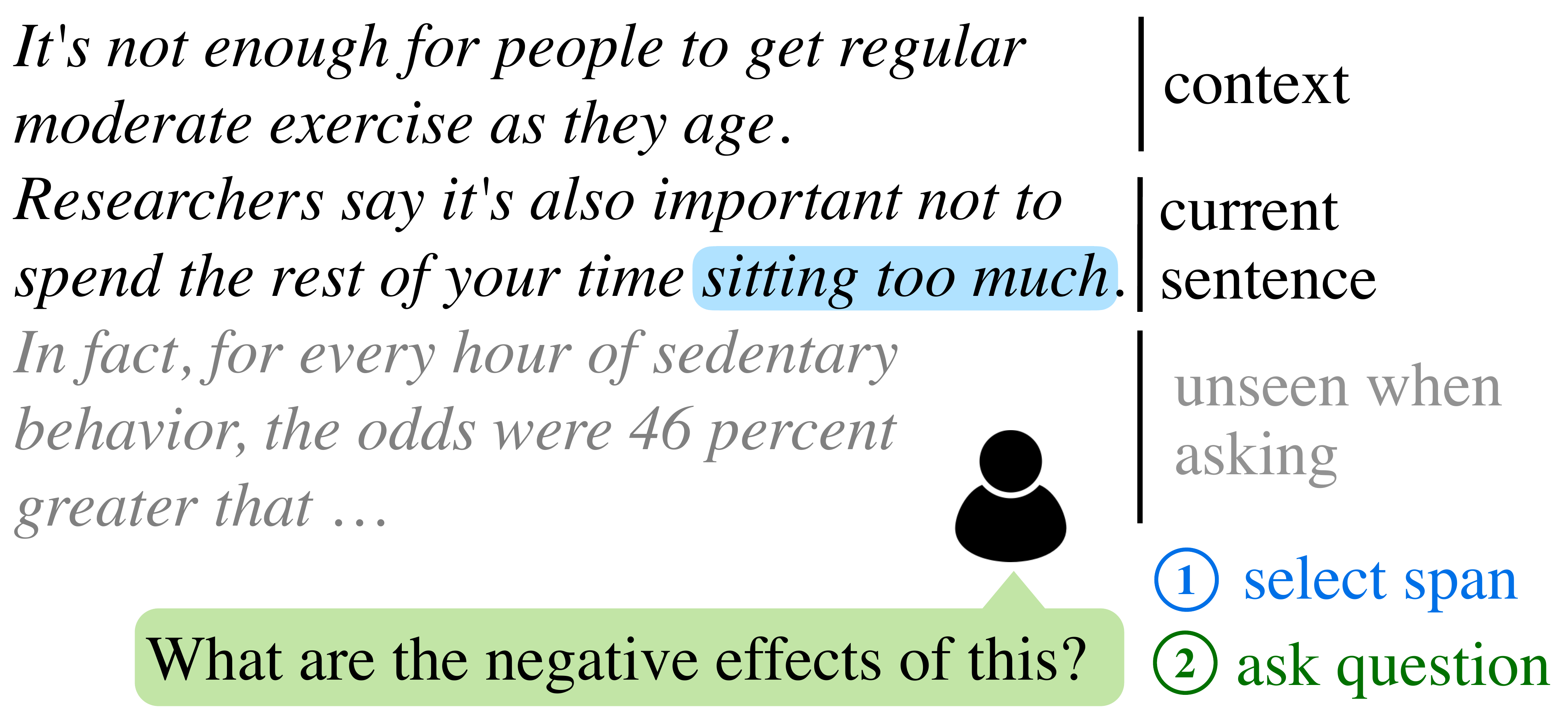}
\end{center}
\caption{Workers highlight spans and ask questions they are curious about the span as they read through the article.}
\label{fig:collection}
\end{figure}
For each sentence, we asked 5 distinct annotators from Amazon Mechanical Turk to ask questions. For quality control, we restrict to workers who are located in English speaking countries, and who have completed at least 100 tasks with an approval rating above 0.98.

\subsection{Question validation}\label{sec:data:validation}

To ensure that our final corpus contain high quality questions, we design a second crowdsourcing task for the validation of these questions, inspired by prior work that also validated crowdsourced questions manually~\cite{QASRL}. At a high level, 
we want to ensure that the questions are grammatically correct and semantically plausible, related to the highlighted span, and not already answered in the sentence or any previous sentence in the article.

Specifically, for each question gathered in Section~\ref{sec:data:turk}, we show the validation annotators the first few sentences of the article up to the sentence where the question is asked, so that the validation annotators have access to the same context as those who asked the questions. We also show the highlighted span for that question. The workers are instructed to answer the following yes/no questions: 
(1) Is it a valid question? An invalid question is incomplete, incomprehensible, not even a question at all, or completely unrelated to the article. 
(2) Is this question related to the highlighted span?
(3) Is the question already answered in prior context? 

Each question is answered by 3 workers from Mechanical Turk. If more than half of the workers judge the question to be either invalid, unrelated to the span, or already answered, the question is deemed low-quality and excluded. About 5\% of the collected questions are low-quality questions; this low rate is consistent with our inspection. Additionally, we manually annotated 100 of the questions removed and we agreed with 92 of them. 

Table~\ref{tab:7} shows the average and standard deviation of the number of tokens for all validated questions, and those of tokens in the highlighted spans. 
 \begin{table}
\centering
\small
\setlength{\tabcolsep}{0.5em}
\begin{tabular}{l|l|l}
   &Average length&std.dev  \\
  \midrule
  Question&7.1 &3.4\\
 Highlighted span&3.2 &2.3\\
\end{tabular}
\caption{Average length and standard deviation of the questions asked by workers and the chosen span}
\label{tab:7}
\end{table}

\subsection{Corpus setup}
For experimentation, in order to ensure model generalizability, we split by articles instead of by sentences: we set aside 50 articles from each news source for validation, which contains 1991 questions, and 50 articles each as the test set, which contains 1894 questions. The remaining articles, with 15931 questions in total, are used as the training set.\footnote{The numbers before filtering are: 2153, 1968, 16816.}
\section{Data analysis}

In this section, we present a deep dive into {\sc Inquisitive}, showing that the questions are much higher level than existing datasets   (Section~\ref{sec:analysis:diversity}), and have rich pragmatic functions (Section~\ref{sec:analysis:pragmatics}). We also investigate the highlighted span associated with each question (Section~\ref{sec:analysis:spans}), and the relative salience of questions to the article  (Section~\ref{sec:analysis:salience}).

\subsection{Question types and diversity}\label{sec:analysis:diversity}

We first investigate the types of questions in the corpus, in comparison to existing question-answering datasets that are also often used in answer-agnostic question generation, in particular, SQuAD 2.0~\cite{SQUAD2} and QuAC~\cite{QuAC}. We additionally compare with NewsQA \cite{newsqa} which also uses news articles.

\paragraph{Question types} To get a basic sense of question types,  Table~\ref{tab:5} shows the most frequent bigrams that start a question. For comparison, we also show the data for SQuAD, QuAC, or NewsQA. It is notable that {\sc Inquisitive} contains a much higher percentage of high level questions --- those signaled by the interrogatives ``why'' and ``how'' --- than either QuAC, SQuAD and NewsQA; these three datasets are characterized by substantially more ``what'' questions. 
 \begin{table*}
\centering
\small
\begin{tabular}{l|l|l|l|l|l|l|l}
  {\sc Inquisitive} & \% &SQuAD& \% &QuAC& \%&NewsQA& \%\\
  \midrule
  what is&8.21&what is&8.49&did he&8.45&what is &8.46\\
  why is&5.73&what was&5.30&what was&8.02&what did &8.38\\
  why did&4.80&how many&4.87&what did&5.40&how many &5.31\\
  what are&3.88&when did&3.13&are there&4.59&who is &4.61\\
  why was&3.54&in what&2.87&how did&3.70&what was &4.41\\
  why are&3.20&what did&2.76&did they&3.39&what does &3.89\\
  how did&3.01&when was&2.14&when did&3.31&who was &2.77\\
  what was&2.20&who was&2.08&what is&3.11&where did &1.79\\
  what does&2.12&what does&1.66&what happened&2.95&what are &1.77\\
  why were&1.88&what are&1.66&what else&2.85&where was &1.59\\
  who is&1.84&what type&1.58&did she&2.12&when did &1.54\\
  why would&1.68&how much&1.10&where did&1.89&where did &1.52\\
  why does&1.62&what year&1.03&what other&1.87&what do &1.46\\
  who are&1.56&where did&1.02&what was&1.74&who did &1.13\\
  how does&1.47&what do&0.86&what were&1.61&what has &1.06\\
\bottomrule
\end{tabular}
\caption{Most frequent leading bigrams in different datasets}
\label{tab:5}
\end{table*}

\paragraph{Lexical diversity}
We also check whether two annotators ask the same questions if they highlighted the same span. 
We estimate this by calculating the average percentages of shared unigram, bigram, and trigrams among the all pairs of questions when the highlighted span exactly match another. 
In the 919 pairs, the percentages of shared ngrams are 33.8 \% for unigrams, 15.4 \% for bigrams, and 8.5\% for trigrams. This shows that even if the annotated spans are exactly the same, annotators often ask different questions.
For example, with the context below (the highlighted span in italic), the annotators asked very different questions: 
\vspace{-0.5em}
\begin{quote}
    \small
    It was the type of weather that would have scrubbed a space shuttle launch. The rain was \emph{relentless}. 
    [\textbf{Q1}: Why was the rain relentless?] 
    [\textbf{Q2}: How heavy is considered relentless?]
\end{quote}
\vspace{-0.5em}
Additionally, we found that the percentage of words appearing in the highlighted span that also appeared in the corresponding question is only 22\%, showing that the annotators are not simply copying from the span they had highlighted into the question.

To estimate the overall lexical diversity of the collected questions, we report the distinct bigram metric~\cite{MMI} that is often used to evaluate lexical diversity of dialog generation systems. This metric calculates the number of distinct bigrams divided by the total number of words in the all questions. 
The metric of our dataset is 0.41; this is much higher than QuAC (0.26), NewsQA (0.29), and slightly higher than SQuAD 2.0 (0.39),\footnote{We use a subset of the same size as our data, taken from the beginning of each dataset, since the metric is sensitive to the amount of data.} possibly because SQuAD contains highly specific questions with many named entities.

\subsection{What information are readers asking for?}\label{sec:analysis:pragmatics}

To gain insights into the pragmatic functions of the questions, we manually analyze a sample of 120 questions from 37 sentences. Of those questions, 113 of them are judged as high-quality using the validation process described in Section~\ref{sec:data:validation}. We now describe a wide range of pragmatic phenomena that signal semantic and discourse understanding of the document. With each category, we also show examples where the highlighted span corresponding to the question are displayed in \emph{italic}. Due to space constraints, we show only the minimal required context to make sense of the questions in the examples.

\paragraph{Why questions}
Causal questions --- those signaled by the interrogative ``why'' as well as its paraphrases such as ``what is the reason that'' --- account for 38.7\% of the questions we inspected. Such why-questions tend to associate very often with an adjective or adverb in the sentence:
\begin{quote}
\vspace{-0.5em}
    \small Predicting the financial results of computer firms has been a \emph{tough job lately}. [\textbf{Q:} Is there a particular cause for this?]
\vspace{-0.5em}
\end{quote}
Verbs also trigger why-questions, indicating that readers are making causal inferences around events:
\begin{quote}
\vspace{-0.5em}
    \small The stock market's \emph{dizzying gyrations} during the past few days have made a lot of individual investors wish they could buy some sort of insurance. [\textbf{Q:} Why is it gyrating?]
\end{quote}

\paragraph{Elaboration questions}
In 21.6\% of the questions, readers seek more detailed descriptions of concepts in the text ranging from entities and events to adjectives. These questions are very often ``how'' questions, although they can take different forms, especially if the question is about an entity:
\begin{quote}
\vspace{-0.3em}
    \small 
    The solution, at least for some investors, may be a \emph{hedging technique} that's well known to players in the stock-options market . 
    [\textbf{Q:} What is the technique?]
    \\
    Undeterred by such words of caution, corporate America is flocking to Moscow, lured by a huge untapped market and Mikhail Gorbachev's \emph{attempt to overhaul the Soviet economy.}
    [\textbf{Q:} How is he overhauling the economy?]
\vspace{-0.5em}
\end{quote}

The second example above shows a elaboration question for a verb phrase; in this case, the patient and the agent are both specified for the verb phrase. In addition, we found that some elaboration questions about events are about missing arguments, especially at the beginning of articles when not enough context has established, e.g.,
\begin{quote}
\vspace{-0.3em}
    \small
    It was the kind of \emph{snubbing} rarely seen within the Congress, let alone within the same party. [\textbf{Q:} Who got snubbed?]
\end{quote}

\noindent\textbf{Definition questions} 
A notable 12.6\% of the questions are readers asking
for the meaning of a technical or domain-specific terminology; for example:
\begin{quote}
\vspace{-0.3em}
    \small
    He said Drexel --- the \emph{leading underwriter} of high-risk junk bonds --- could no longer afford to sell any junk offerings. [\textbf{Q:} What is a leading underwriter?]
\vspace{-0.5em}
\end{quote}
In some cases, a simple dictionary or Wikipedia lookup will not suffice, as the definition sought can be highly contextualized:
\begin{quote}
\vspace{-0.3em}
    \small
    Mrs.~Coleman, 73, who declined to be interviewed, is the Maidenform \emph{strategist}. [\textbf{Q:} What is the role of a strategist?]
\vspace{-0.5em}
\end{quote}
Here the role of a strategist depends on the company that employed her.

\paragraph{Background information questions} 
We found that 10\% of the questions aim at learning more about the larger picture of the story context, e.g., when the author draws comparisons with the past:
\begin{quote}
\vspace{-0.3em}
    \small Seldom have \emph{House hearings caused so much apprehension} in the Senate. [\textbf{Q:} When have house hearings caused apprehension?]
\vspace{-0.5em}
\end{quote}
Other times, answering those questions will provide topical background knowledge needed to better understand the article:
\begin{quote}
\vspace{-0.3em}
    \small The stock market's dizzying gyrations during the past few days have made a lot of individual investors wish they could buy some sort of \emph{insurance}. [\textbf{Q:} How would that insurance work?]
\end{quote}

\paragraph{Instantiation questions}
In 8.1\% of the questions, the readers ask for a specific example or instance when a set of entities is mentioned; i.e., answering these questions will lead to entity instantiations in the text~\cite{McKinlay}:
\begin{quote}
\vspace{-0.3em}
    \small
    The solution, at least for \emph{some investors}, may be a hedging technique that's well known to players in the stock-options market. [\textbf{Q:} Which ones?]
\vspace{-0.5em}
\end{quote}
This indicates that the reader sometimes would like concrete and specific information.

\paragraph{Forward looking questions}
Some questions (4.5\%) reflect that the readers are wondering ``what happened next'', i.e., these questions can bear a reader's inference on future events:
\begin{quote}
\vspace{-0.3em}
    \small
    Ralph Brown was 31,000 feet over Minnesota when both jets on his Falcon 20 flamed out. At 18,000 feet, he says, he and his co-pilot ``were looking for an \emph{interstate} or a cornfield'' to land.
    [\textbf{Q:} Would they crash into cars?]
\end{quote}

\paragraph{Others}
We have noticed several other types of questions, including asking about the specific timeframe of the  article (since readers were only shown the body of the text). Some readers also asked rhetorical questions, e.g., expressing surprise by asking \emph{Are they really?} to an event.

Finally, we observed that a small percentage of the questions are subjective, in the sense that they reflect the reader's view of the world based on larger themes they question, e.g., \emph{Why is doing business in another country instead of America such a sought-after goal?}.

\begin{table}
\centering
\small
\setlength{\tabcolsep}{0.5em}
\begin{tabular}{ll|ll}
   Constituent & \% & Constituent & \%  \\
  \midrule
  NP&22.5 &VBN&2.8 \\
  NN&13.5 & VBD&2.7 \\
  JJ&9.6 &  ADJP&2.5\\
  VP&4.9 & VB&2.2 \\
  NNS&4.3 & S&2.0 \\
  NNP&3.9 & VBG&1.9 \\
  NML&2.8 & PP&1.4\\
\end{tabular}
\caption{Top constituents in highlighted spans.}
\label{tab:10}
\end{table}

\subsection{What do readers ask about?}\label{sec:analysis:spans}

In addition to understanding what type of information is sought after, we also investigate whether there are regularities in what the questions are about. This is reflected in the highlighted spans accompanied with each question.

Table~\ref{tab:10} shows the most frequent syntactic constituents that are highlighted.\footnote{Parsed by Stanford CoreNLP  \cite{corenlp}.} Readers tend to select short phrases (the average number of tokens in the spans is 3.2) or individual words; while noun phrases are most frequently selected, a variety of constituents are also major players.

We found that the probability that two highlighted spans overlap is fairly high: 0.6. However, F1-measure across all spans pairs is only 0.25. The percentages of highlighted tokens chosen by 2, 3, 4, or all 5 annotators are: 0.8, 0.17, 0.03, and 0.006.
Upon manual inspection, we confirm the numerical findings that there is a high variance in the location of highlights, even though the question quality is high. Secondly, while there are spans that overlap, often a question's ``aboutness'' can have many equally valid spans, especially within the same phrase or clause. This is exacerbated by the short average length of the highlights.

\subsection{Question salience}\label{sec:analysis:salience}
The analysis in Section~\ref{sec:analysis:pragmatics} implied that the information readers sought after differs in terms of their relative salience with respect to the article: some information (e.g., background knowledge) can be important but typically isn't stated in the article, while others (e.g.,  causal inference or elaboration) are more likely to be addressed by the article itself. To characterize this type of salience in our data, we ask workers to judge if the answer to each question \emph{should} be in the remainder of the article, using a scale from 1 to 5.
Each validated question is annotated by 3 workers.
The average salience rating is 3.18, with a standard deviation of 0.94, showing that the questions are reasonably salient. The distribution of salience ratings is shown in Table~\ref{tab:8}.

 \begin{table}[t]
\centering
\small
\setlength{\tabcolsep}{0.5em}
\begin{tabular}{ll|ll}
   Range&\% & Range&\%  \\
  \midrule
 $[1,2)$&6.9 & $[3,4)$&41.6 \\
 $[2,3)$&22.5 & $[4,5]$&29.1 \\
\end{tabular}
\caption{Distribution of question salience ratings.}
\label{tab:8}
\end{table}

\section{Question generation from known spans}\label{sec:exp:qg}

\begin{table*}
\centering
\small
\setlength{\tabcolsep}{0.5em}
\begin{tabular}{l|llllll|l}
  Model & Train-2 & Train-3 & Train-4& Article-1& Article-2& Article-3 & Span\\
  \midrule
  Ours & 0.627 &0.352& 0.135&0.397&0.147&0.0877 &0.278\\
  Ours + SQuAD pretraining& 0.603&0.340 & 0.145&0.404&0.155&0.0931&0.284 \\
  Human & 0.520&0.229 & 0.069&0.341&0.115&0.064&0.219 \\
\bottomrule
\end{tabular}
\caption{$n$-gram overlap between the generated question and either the training set, the conditioned span, and the source article. We see that the model generates novel questions without substantial copying from any of these sources, approaching the novelty rates in human questions.}
\label{tab:1}
\end{table*}

We use {\sc Inquisitive} to train question generation models and evaluate the models' ability to generate questions \emph{with access to} the gold-standard spans highlighted by annotators, while also contrasting this task with question generation from SQuAD 2.0 data. In Section~\ref{sec:exp:pipeline}, we present experiments that simulate the practical scenario where the highlighted spans are not available. 

\subsection{Models}

\begin{figure}
\begin{center}
\includegraphics[width=0.48\textwidth]{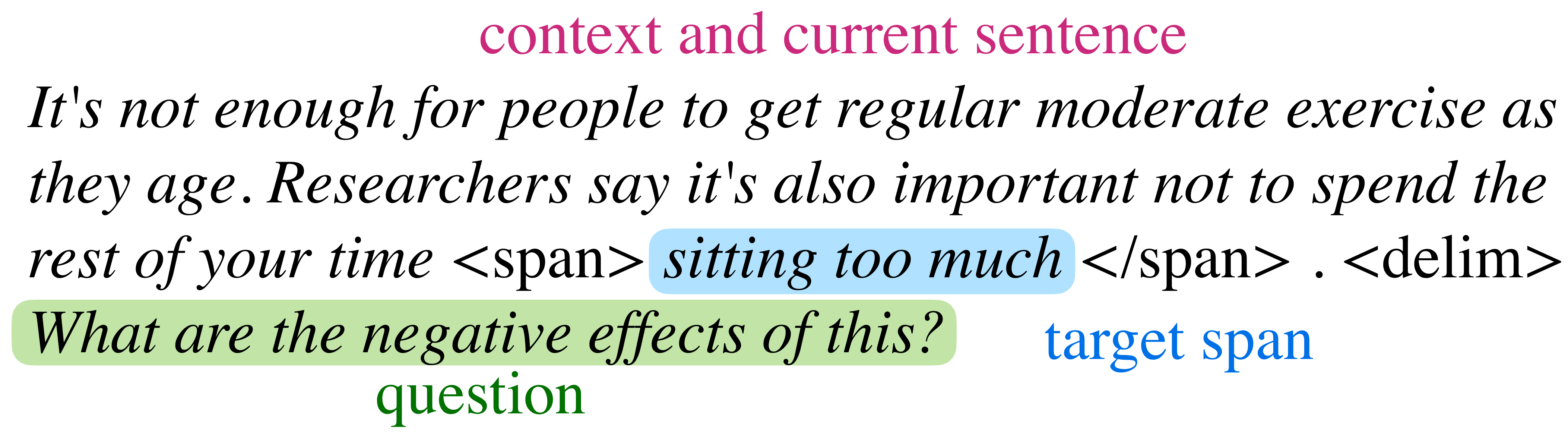}
\end{center}
\caption{We concatenate the context, sentence and questions and learn a language model on them using GPT-2.}
\label{fig:gpt2}
\end{figure}

Our question generation model is based on GPT-2~\cite{gpt2}, a large-scale pre-trained language model, which we fine-tune on our data. Each training example consists of two parts, illustrated in Figure \ref{fig:gpt2}. The first part is the article context, from the beginning to the sentence where the question is asked. Two special tokens are placed in line to indicate the start and end positions of the  highlighted span. The second part is the question. The two parts are concatenated and separated by a delimiter. The model is trained the same way as the GPT-2 language modeling task, with the loss only 
accumulated for the tokens in the question. During testing, we feed in the article context and the delimiter, and let the model continue to generate the question. We call this model, which uses the span and all prior context, \textbf{Inquirer}.

To analyze the contribution of prior context that serve as common ground when humans generated the questions, we train two additional variants of the model: 
(1) A \textbf{span+sentence} model, in which the conditioning context only contains the single sentence where the question is asked. (2) A \textbf{span}-only model, in which the conditioning context only contains the highlighted span inside the sentence where the question is asked. 

\paragraph{SQuAD pre-training.}
To investigate whether models could leverage additional supervision from SQuAD,
we experiment on our Inquirer model that is first fine-tuned on SQuAD 2.0. 
We treat the SQuAD answer spans as the highlighted spans. Note that in SQuAD, the span is where the answer to the generated question should be; in our task, the span is what the question should be \emph{about}. This parallel format allows us to show the pragmatic distinction of {\sc Inquisitive} question generation  and SQuAD-based question generation. Nevertheless, such pre-training could in principle help our model learn a language model over questions, albeit ones with a different distribution than our target data (Table \ref{tab:5}).

\paragraph{Model parameters.}
We use the GPT2-medium model for all question generation models. The batch size is 2 and we fine-tune for 7 epochs. For SQuAD pre-training, the batch size is 1 and we fine-tune for 1 epoch.
We use the Adam \cite{adam} optimizer with $(\beta_1,\beta_2)=(0.9,0.999)$ and a learning rate of 5e-5. The parameters are tuned by manually inspecting generated questions in the validation set.

\subsection{Human evaluation}
We mainly perform human evaluation on the generated questions by collecting the same validity judgments from Mechanical Turk as described in Section~\ref{sec:data:validation}. 

The results are shown in Table \ref{tab:results-human}. We can see that \emph{Inquirer} generates a valid question 75.8\% of the time, the question is related to the span 88.7\% of the time, and the question is already answered 9.6\% of the time. This shows that \emph{Inquirer} is able to learn to generate reasonable questions. Compared to crowdsourcing workers, \emph{Inquirer} questions are as related to the given span and more salient, though the questions are more often invalid or already answered.

With SQuAD pre-training, human evaluation results did not improve, showing that the structure of questions learned from SQuAD is not enough to offset the difference between the two tasks.

The results also show that removing context makes the questions less valid and related to the span. The weakest model is the span-only one, where no context is given but the span. This finding illustrates the importance of building up common ground for the question generation model, illustrated in an example output below:
\begin{quote}
\small
    \textbf{Context:} Health officials said Friday they are investigating why a fungicide that should have been cleaned off was found on apples imported from the United States . In a random sampling of apples purchased at shops in the Tokyo area , two apples imported from Washington State were found to have trace amounts of the fungicide , health officials said . " This is not a safety issue by any means , " said U . S . embassy spokesman Bill Morgan . " It ' s a technical one . This fungicide is also commonly used by farmers in Japan . " Several stores in Tokyo that stocked apples packed by Apple King , a packer in Yakimo , Washington state , were voluntarily recalling the apples Friday because of possible health hazards , a city official said . \\
    \textbf{Sentence:} But a spokesman for Japan ' s largest supermarket chain , Daiei Inc . , said the company has \emph{ no plans to remove U . S . apples from its shelves} .\\
    \textbf{Inquirer:} If this is the case, why have no plans to remove them from their shelves? \\
    \textbf{Span+sentence}: Why would the supermarket chain remove U.S. apples from its shelves?
\end{quote}

\subsection{Automatic evaluation}
We also use several metrics that capture the generation behavior of the models.
These include: (1) Measuring the extent of copying. \textbf{Train-n}: percentage of n-grams in the generated questions that also appeared in some question in the training set. This metric could show if the model is synthesizing new questions or simply copying questions from the training set.  \textbf{Article-n}: percentage of n-grams in the generated questions that also appeared in either the sentence where the question is asked, or any prior sentence in the same article. This metric estimates the extent to which the system is copying the article. 
(2) \textbf{Span}: percentage of words appeared in the annotated span that also appeared in the generated question. This is an rough estimation of how the question is related to the span. 
Table \ref{tab:1} shows that \emph{Inquirer}-generated questions are not simply copied from the training data or the article, though the n-gram scores are comparatively much higher than questions asked by human.

\section{Question generation from scratch}\label{sec:exp:pipeline}
In this section, we assume that spans are not given, and discuss two approaches to generating questions from scratch: a pipeline approach with span prediction, and a question generation model without access to any span information.

\subsection{Models}

\noindent\textbf{Pipeline.} The pipeline approach consists of 2 stages: span prediction and question generation using Inquirer. To predict the span, we use a model similar to the BERT model for the question answering task~\cite{BERT} on SQuAD 1.1~\cite{SQUAD}. We replace the concatenation passage and question with a concatenation of the target sentence and the previous sentences in the document. Now, the span to ask a question about is treated analogously to the answer span in SQuAD: we find its position in the ``passage'' which is now the target sentence.\footnote{We use the pretrained bert-large-uncased-whole-word-masking model, and fine-tune it for 4 epochs. The learning rate is 3e-5, batch size 3, maximum sequence length 384. We use the Adam optimizer with $(\beta_1,\beta_2)=(0.9,0.999)$. The parameters are tuned by manually inspecting generated questions in the validation set.}

\noindent\textbf{Sentence+Context.} We also experiment with a \emph{Sentence+Context} model, where the model has no knowledge of any span information in both training and testing; it is trained based purely on (context, sentence) pairs from our dataset.
This baseline evaluates the usefulness of predicting the span as an intermediate task.

\subsection{Results}

\paragraph{Question evaluation} 
Human evaluation results for \emph{Inquirer-pipeline} and \emph{Sentence+Context} are shown in Table~\ref{tab:results-human}. 
\emph{Inquirer-pipeline} is able to produce questions with a validity performance only slightly below \emph{Inquirer}. However, without access to gold spans, more questions are already answered, and are unrelated to the predicted spans.  
Yet predicting the spans is clearly useful: 
compared with \emph{Sentence+Context}, which does not use the spans at all, \emph{Inquirer-pipeline} generates more valid questions. While more of these are already answered in the text, we argue that it is more important to ensure that the questions make sense in the first place, 
illustrated in this example below:
\begin{quote}
    \small
    \textbf{Context:} Bank of New England Corp . , seeking to streamline its business after a year of weak earnings and mounting loan problems , said it will sell some operations and lay off 4 \% of its work force . The bank holding company also reported that third - quarter profit dropped 41 \% , to \$ 42 . 7 million , or 61 cents a share , from the year - earlier \$ 72 . 3 million , or \$ 1 . 04 a share . \\
    \textbf{Sentence:} Among its restructuring measures , the company said it plans to sell 53 of its 453 branch offices and to lay off 800 employees .\\
    \textbf{Inquire Pipeline:} What other measures did the company have?\\
    \textbf{Sentence+context:} What are those?
\end{quote}

\paragraph{Span evaluation}
Finally, we present the results for the intermediate task of span prediction. Note however that since we already observed that spans can be highly subjective (c.f.\ Section~\ref{sec:analysis:spans}), these results should be interpreted relatively.

We use two metrics: (1)  \emph{Exact match}: the percentage of predictions that exactly match one of the gold spans in a sentence. (2) \emph{Precision}: the portion of tokens in the predicted span that is also in a gold span. 
Since there are usually multiple spans annotated in a sentence, we report the micro-average across all spans.
We do not report recall or F1, since considering recall would lead to a misleading metric that prefers long span predictions. In particular, the ``best'' length that optimizes F1 is about 20 tokens, while the average length of gold spans is 3.2 tokens.

Table \ref{tab:2} shows the results of span prediction. 
Our span prediction model is compared with 3 scenarios. The Random model picks a span at a random position with a fixed length that is tuned on the validation set. The human-single scores are the average scores between two spans that are highlighted by different annotators. The human-aggregate baseline compares the annotation of one worker against the aggregation of the annotations of the other 4 workers. 
These results show that the task is highly subjective, yet our model appears to agree with humans at least as well as other humans do.

\section{Conclusion}

We present {\sc Inquisitive}, a large dataset of questions that reflect semantic and discourse processes during text comprehension. We show that people use rich language and adopt a range of pragmatic strategies to generate such questions. We then present  question generation models trained on this data, demonstrating several aspects that generating {\sc Inquisitive} questions is a feasible yet challenging task.

\section*{Acknowledgements}
We thank Shrey Desai for his comments, and the anonymous reviewers for their helpful feedback. We are grateful to family and friends who supported the authors personally during the COVID-19 pandemic. This work was supported by NSF Grant IIS-1850153.

\begin{table}
\centering
\small
\setlength{\tabcolsep}{0.5em}
\begin{tabular}{l|llll}
  Model & Valid & Related & Answered & Salience\\
  \midrule
  \multicolumn{5}{c}{Conditioning on gold span}\\
  \midrule
  Span  & 0.592 & 0.746 & 0.082  & 3.12\\
  Span+Sentence & 0.719&0.867 & 0.086&3.64 \\
  Inquirer & 0.758 &0.887& 0.096 &3.65\\
  Inquirer+SQuAD & 0.742&0.854 & 0.075&3.59 \\
  \midrule
  \multicolumn{5}{c}{From scratch}\\
  \midrule
  Sentence+Context&0.711&-& 0.115 & 3.01  \\   
  Inquirer Pipeline &0.748&0.777& 0.178 & 3.00  \\  Human&0.958&0.866& 0.047 & 3.18  \\
\bottomrule
\end{tabular}
\caption{Human evaluation results for generated questions. Conditioning on the immediate sentence (+Sentence) and further context (+Context) help generate better questions, but SQuAD pre-training does not.}
\label{tab:results-human}
\end{table}

\begin{table}[t!]
\centering
\small
\setlength{\tabcolsep}{0.5em}
\begin{tabular}{l|llll}
  Model & Exact & Precision\\
  \midrule
  Ours &  0.121&0.309\\
  Random&0.002 &0.118  \\
  Human-single& 0.075&0.265   \\
  Human-aggregate  & 0.115 &0.391 \\
\bottomrule
\end{tabular}
\caption{Results for predicting the span. Note that human agreement is low, so our automatic method is on par with held-out human comparisons.}
\label{tab:2}
\end{table}

\bibliography{anthology,emnlp2020}
\bibliographystyle{acl_natbib}

\newpage

\appendix

\section{Crowdsourcing Instructions}
\begin{figure}[h!]
\begin{center}
\includegraphics[width=1.0\textwidth]{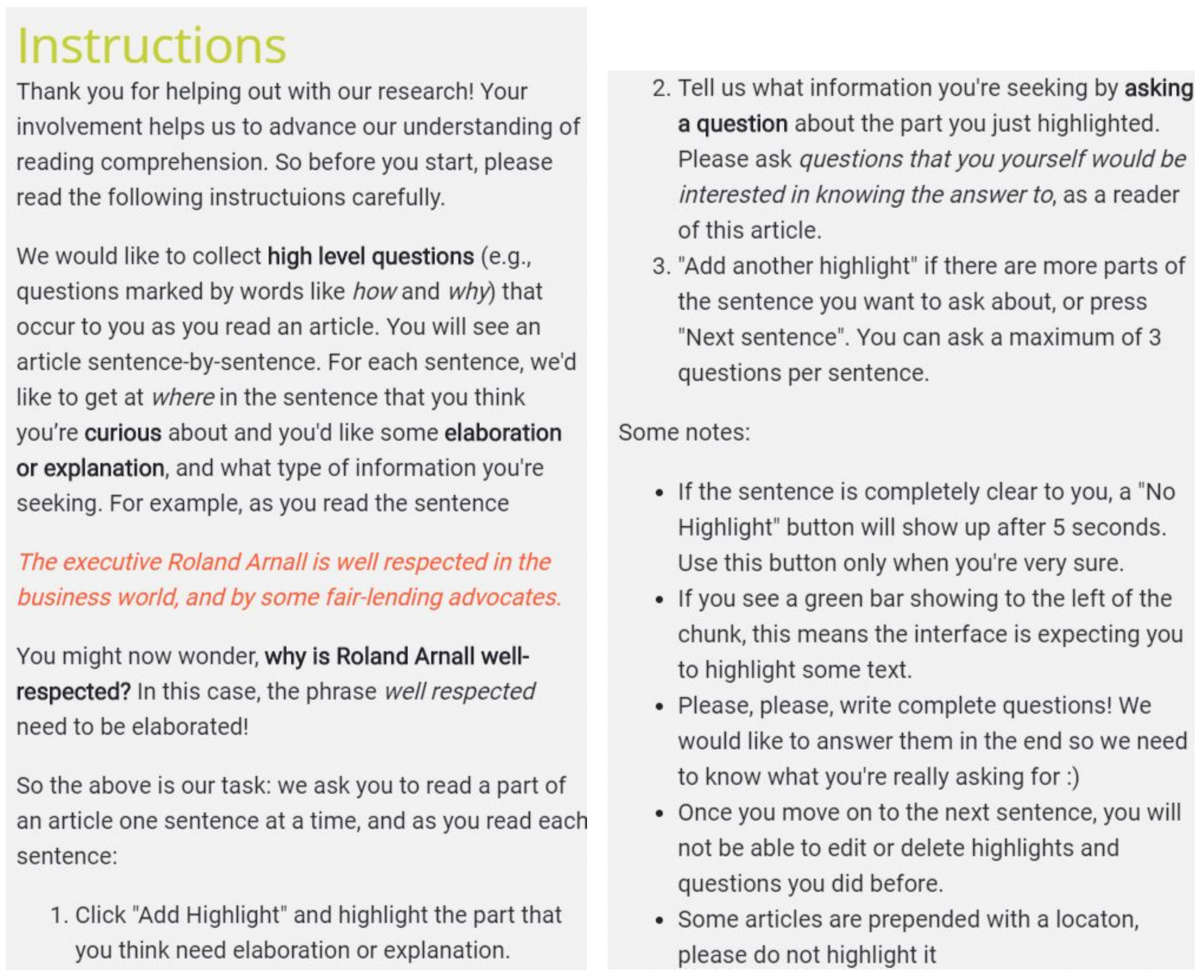}
\caption{Instructions for question collection.}
\end{center}
\label{fig:Instructions}
\end{figure}

\begin{figure}[h!]
\begin{center}
\includegraphics[width=1.0\textwidth]{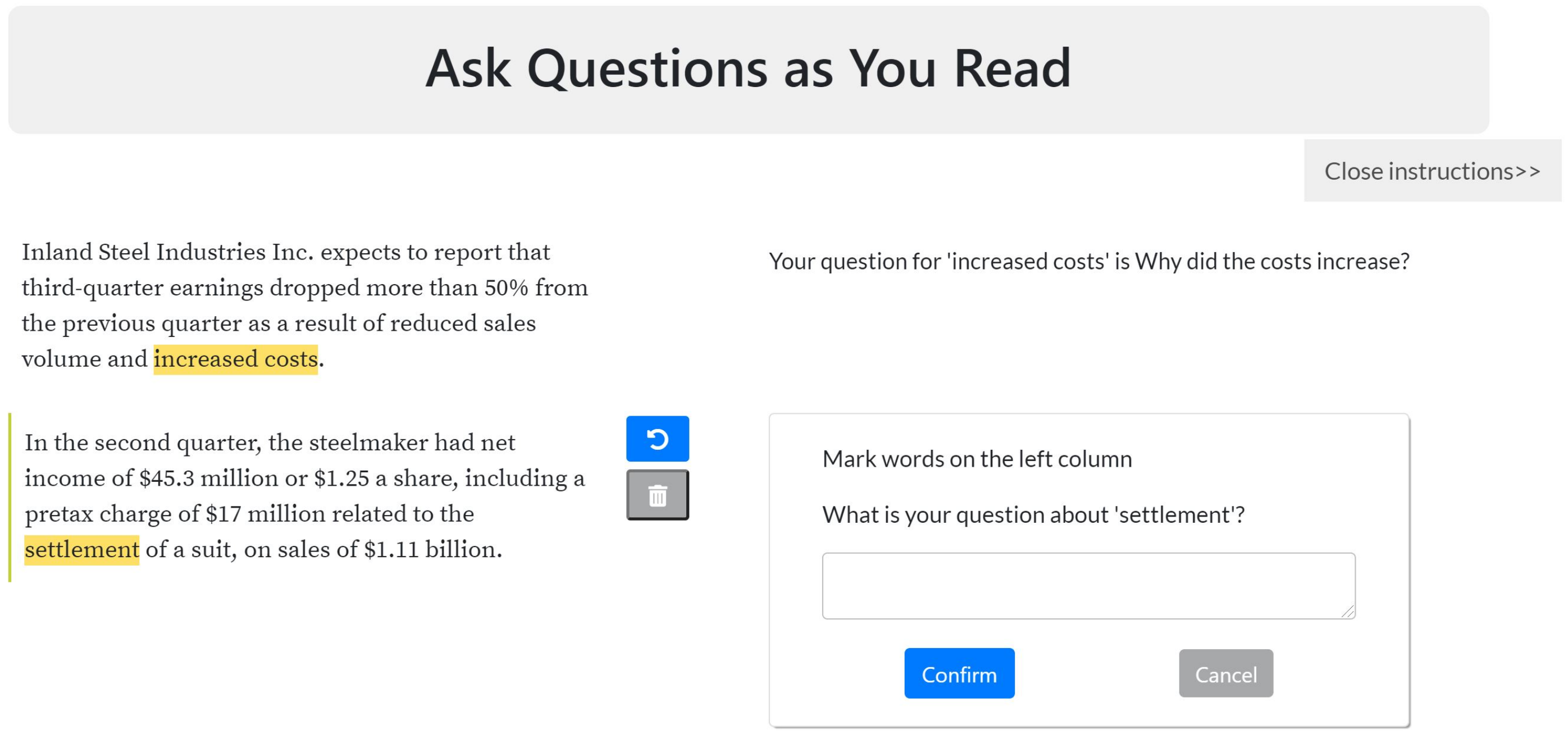}
\caption{Turk interface for collecting questions.}
\end{center}
\label{fig:interface}
\end{figure}

\end{document}